# SHAP values for Explaining CNN-based Text Classification Models

Wei Zhao[1], Tarun Joshi, Vijayan N. Nair, and Agus Sudjianto

Corporate Model Risk, Wells Fargo, USA

August 19, 2020

**Abstract**

Deep neural networks are increasingly used in natural language processing (NLP) models. However, the need to interpret and explain the results from complex algorithms are limiting their widespread adoption in regulated industries such as banking. There has been recent work on interpretability of machine learning algorithms with structured data. But there are only limited techniques for NLP applications where the problem is more challenging due to the size of the vocabulary, high-dimensional nature, and the need to consider textual coherence and language structure. This paper develops a methodology to compute SHAP values for local explainability of CNN-based text classification models. The approach is also extended to compute global scores to assess the importance of features. The results a[1]re illustrated on sentiment analysis of Amazon Electronic Review data.

## 1. Introduction

Advances in machine learning (ML), particularly deep neural networks (DNNs), have allowed us to extract and use the extensive information in unstructured data for applications in public and private sectors. In the banking area, these include modeling and analysis of text data on customer communications, complaints, emails, etc. However, the complexity of the underlying ML algorithms makes the results hard to interpret and explain to users and regulators.

Interpretability of ML algorithms has been the subject of considerable recent research, and several useful techniques have been developed for structured data. Global diagnostics for identifying important features include variable-importance analysis (Friedman, 2001), Sobol indices for sensitivity analysis (Kucherenko S. T., 2012) (Chen, Hu, Nair, & Agus, 2018), Global Shapely effects (Song, Nelson, & Staum, 2016), derivative-based sensitivity (Kucherenko & others, 2010), and so on. There are also local diagnostics for explaining the behavior of a model at individual observations or in a local region: LIME (Ribeiro, Singh, & Guestrin, 2016), Leave One Covariate Out (LOCO) (Lei, G'Sell, Rinaldo, Tibshirani, & Wasserman, 2018), SHAP explanation (Lundberg & Lee, 2017; Lundberg, Erion, & Lee, 2018), Quantitative input influence (Datta, Sen, & Zick, 2016), etc. Additional techniques for neural networks include Integrated Gradients (Sundararajan, Taly, & Yan, 2017), DeepLIFT (Shrikumar, Greenside, & Kundaje, 2017), Layer-wise Relevance Propagation (LRP) (Binder, Montavon, Lapuschkin, Müller, & Samek, 2016), and Derivative based sensitivity analysis (Liu, Chen, Nair, & Sudjianto, 2019).

---

[1] Email: vivienzhao0119@hotmail.com

However, many of these methods cannot be used with, or easily generalized to, unstructured data. Further, the nature of unstructured data raises additional challenges: i) the vocabulary size in some applications; ii) the high-dimensional nature of the problems; and iii) the need to retain textual coherence and language structure. Methods developed so far have focused on simple models that use bag-of-words and provide explanations in terms of uni-grams/words.

In this paper, we develop a method to use SHAP values for local explainability with text classification models based on convolutional neural networks (CNNs). Text classification is an important subclass of problems in natural language processing (NLP). It is used in sentiment analysis, complaints identification, fraud detection, and so on. We focus on CNN as it has been shown to have good performance in text classification tasks and outperforms other ML algorithms such as support vector machines (SVM) and Gradient Boosting Machines (GBM). Limited research suggests that it may perform better than other deep learning algorithms such as RNN and LSTM. In addition, CNN models can be used as a surrogate to approximate the results from other ML algorithms and potentially be used to explain any text classification model. Therefore, providing useful guidance on explaining CNN models for text classification tasks will be very valuable. As we will see later, CNN with max-pooling learns a fixed-dimension representation of a text document where the dimensions can be mapped back to n-grams in the text. This allows us to explain classification results in terms of interpretable n-grams.

The paper is organized as follows. In Section 2, we review the existing explanation methods for structured data. Then, we discuss the feasibility of extending them to unstructured text data and point out the limitations and challenges. In Section 3, we propose the novel CNN SHAP method for explaining text classification model. We first introduce the kernel SHAP diagnostic and describe how it can approximate exact SHAP values. Then, we introduce the CNN model and explain how it can reduce the explanation dimension. Then, we describe the detailed steps in applying CNN SHAP for local explanation and use the Amazon Electronic Reviews Sentiment Analysis model to demonstrate our results. We extend it to global explanation by introducing global ranking scores and illustrate the approach on the Amazon dataset. Finally, Section 5 provides concluding remarks and potential future research directions. The Appendix provides a limited study to quantify information in subsampling the weights in computing kernel SHAP using weighted least squares.

## 2. Existing Explanation Methods

This section reviews existing methods for interpreting black-box ML models. Section 2.1 focuses on methods for sturctured data. Section 2.2 discusses whether these methods can be extended easily to text data. In particular, we point out the challenges in NLP model explanation.

## 2.1 Explaining Structured Data

Techniques for explaining ML prediction models have been recently proposed and studied for structured data. Generally, we can group them into three categories:

1) Surrogate model based diagnostics;
2) Shapley values based diagnostics; and
3) Gradient based diagnostics

The concept behind diagnostics based on surrogate models is to approximate the complex structure of ML models by simpler interpretable models such as linear models. The Local Interpretable Model-agnostic Explanation (LIME) is a local surrogate diagnostic and has been commonly used. Given a point of interest in the predictor space, say $x^*$, it takes a suitable neighborhood of points around $x^*$ and fits a simple local model that is interpretable. Global surrogate methods have also been developed. For example, Surrogate Locally-Interpretable Model or SLIM (Hu, Chen, Nair, & Sudjianto, 2020) uses model-based trees to partition the predictor space and fits generalized-additive main-effects models at each leaf node. It turns out that SLIM generally has good predictive performance as a global model.

SHAP, proposed by Lundberg and Lee (2017), is a local diagnostic that partitions each prediction into individual contributions of the features. To motivate the development of SHAP, let us start with a simple linear regression problem with structured data where the response is continuous. The predictions can be written as:

$$\hat{y}_i = b_0 + b_1 x_{1i} + \cdots + b_d x_{di},$$

where $\hat{y}_i$ is the $i$-th predicted response, $\{x_{1i}, \ldots, x_{di}\}$ are the corresponding predictors, and $\{b_0, \ldots, b_d\}$ are the estimated regression coefficients. If the predictors are independent, the contribution of the $k$-th predictor to the predicted response $\hat{y}_i$ can be unambiguously expressed as $b_k x_{ki}$ for $k = 1, \ldots d$. SHAP is a generalization of this concept to more complex supervised learning models. To do this, define the following:

- $F$ is the entire set of features, and $S$ denotes a subset.
- $S \cup i$ is the union of the subset $S$ and feature $i$.
- $E[f(X)|X_S = x_S]$ is the conditional expectation of model $f(\cdot)$ when a subset $S$ of features are fixed at the local point $x$.

(Lundberg, Erion, & Lee, 2018) define the SHAP value to measure the contribution of the $i$-th feature as

$$\phi_i = \sum_{S \subseteq F \setminus \{i\}} \frac{|S|!(|F| - |S| - 1)!}{|F|!} \{E[f(X)|X_{S \cup i} = x_{S \cup i}] - E[f(X)|X_S = x_S]\}. \quad (1)$$

SHAP has been shown to satisfy good properties such as fairness and consistency on attributing importance scores to each feature. But the calculation of SHAP scores is computationally expensive (Lundberg, Erion, & Lee, 2018). We will describe the efficient model-agnostic kernel SHAP approximation algorithm (Lundberg & Lee, 2017) in Section 3.

The above methods try to explain the model from outside of the 'box', looking only at what is fed into the model (i.e., model input $x$) and what is produced (i.e., model output $f(x)$). Gradient-based algorithms such as Integrated Gradients, DeepLIFT, and Layer-wise Relevance Propagation look inside the 'box' and use the gradient information. For example, they take the product of the gradient and feature values as a reasonable starting point for attributing contributions. They have some advantages such as fast computation, especially for Neural Networks where the gradients are available.

### 2.2 Explaining Unstructured Data

There has not been as much work in interpreting unstructured data related tasks such as sentiment analysis and document classification. Some of the methods discussed in Section 2.1 can be extended to text data after we convert it into numerical data, for example, bag-of-word (BOW) or TF-IDF representations, and treat individual word as a feature of a text document.

But the computations of SHAP values and other techniques discussed in Section 2.2 is challenging with NLP tasks due to the high-dimensionality of the input features (large vocabulary size). In fact, the computational complexity increases exponentially and computing exact SHAP values becomes intractable for feature dimensions of 100 or higher.

Document representation based on word embedding is a common method in text data related tasks and often provides satisfactory results. However, this leads to an explanation in terms of embedding dimensions instead of text. The dimensions are not typically interpretable and hence the diagnostics based on the embedded document is confusing. This is a common issue for all diagnostics like LIME and SHAP, where the output importance scores are corresponding to the embedding columns, instead of interpretable words or phrases. In general, we need to map the scores of the dimensions back to words, which typically means projecting from low dimension space to a high dimension space. Currently, there is no theoretical support for that.

## 3. CNN SHAP: Explaining CNN Text Classification Model Using Kernel SHAP

In this section, we propose a novel approach to compute SHAP diagnostic for CNN-based Text Classification models. We generalize the application of kernel SHAP to unstructured data utilizing the structure of CNN models. CNNs play the role of a dimension-reduction feature extractor, transforming the original high-dimensional unstructured text space into structured low-dimensional feature space. It uses the convolutional layer and max-pool layer as filters that select n-grams from the text, so we are able to identify the n-grams each filter selected. Then SHAP is applied over the classification layer which is constructed over the max-pool layer outputs. We will show how n-grams picked up by CNN model contribute to the final predictions.

Before we go into details, we first introduce the kernel SHAP technique and the structure of CNN. The following terms are used in the rest of this paper:

- Filter: convolutional layer + activation function + max-pool layer
- Filter size: the kernel size of the 1D convolution, which directly determines the size of n-grams fed into the filter.
- Extracted feature: the max-pool layer outputs.
- Classification layer: the feed-forward neural network constructed over the max-pool layer outputs.

## 3.1 Basic Structure of a 1D Text Convolutional Neural Network

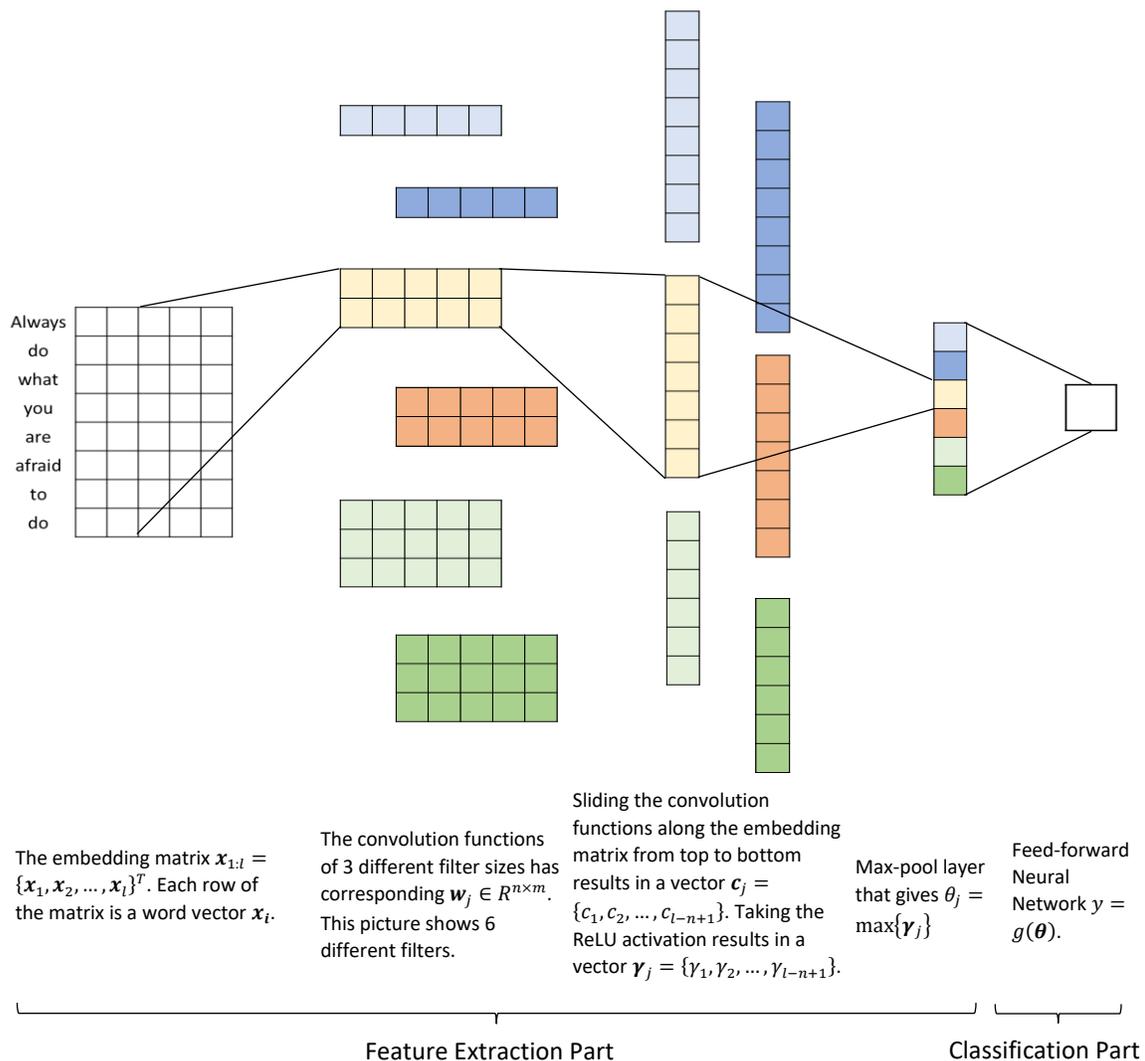

The embedding matrix $x_{1:l} = \{x_1, x_2, \dots, x_l\}^T$. Each row of the matrix is a word vector $x_i$.

The convolution functions of 3 different filter sizes has corresponding $w_j \in R^{n \times m}$. This picture shows 6 different filters.

Sliding the convolution functions along the embedding matrix from top to bottom results in a vector $c_j = \{c_1, c_2, \dots, c_{l-n+1}\}$. Taking the ReLU activation results in a vector $\gamma_j = \{\gamma_1, \gamma_2, \dots, \gamma_{l-n+1}\}$.

Max-pool layer that gives $\theta_j = \max\{\gamma_j\}$

Feed-forward Neural Network $y = g(\theta)$.

Feature Extraction Part | Classification Part

*Figure 1 Structure of an Example CNN Text Classification Model. The model is consist of 4 layers: 1. A embedding layer that embedded input texts to high dimensional space; 2. A convolutional layer in different sizes formed by convolution functions and ReLU function; 3. A max-pooling layer takes the maximum value of convolution outputs; 4. Feed-forward neural network that gives the final outputs of model.*

Figure 1 provides a graphical view of the structure of a CNN Text Classification algorithm. Let

- $l$ be the total words in the text input,
- $m$ be the embedding dimension, and
- $n$ be the filter size, which means the filter is an $n$-gram filter,
- $h$ be the total number of filters for all filter size.

Suppose each word in the vocabulary is embedded into an $m$-dimensional vector space, and let $x_i \in R^m$ denote the word embedding. For a text input with $l$ words, the embedding layer is then an $l \times m$ matrix denoted as

$$x_{1:l} = \{x_1, x_2, \dots, x_l\}^T \in R^{l \times m}.$$

With a size $n$ filter $f_j \in R^{n \times m} \to R, j \in \{1,2,\dots,h\}$ (convolution function) sliding over the embedding layer, we get

$$c_k = f_j(x_{k:k+n-1}) = w_j \cdot x_{k:k+n-1} + b_j,$$

where $k \in \{1,2,\dots,l-n+1\}$, and $w_j \in R^{n \times m}$,

$$\text{and } c_j = \{c_1, c_2, \dots, c_{l-n+1}\}.$$

The ReLU activation is applied over the convolutional output

$$\gamma_j = \{\gamma_1, \gamma_2, \dots, \gamma_{l-n+1}\},$$

where $\gamma_k = \max(0, c_k)$. Then a max-pool layer picks up the maximum value from the ReLU output and gets

$$\theta_j = \max\{\gamma_j\} = \max\{\gamma_1, \gamma_2, \dots, \gamma_{l-n+1}\}.$$

Finally, a feed-forward neural network is applied to the max-pool layer output to get

$$y = g(\theta),$$

where $\theta = \{\theta_1, \theta_2, \dots, \theta_h\}$ is the max-pool layer output from all $h$ convolutional filters.

### 3.2 Kernel SHAP

Kernel SHAP (KS) is a model-agnostic method proposed by Lundberg and Lee (2017) to approximate the SHAP values. Consider the terms (or weights) in front of the conditional expectations in equation (1):

$$w_S = \frac{|F| - 1}{\binom{|F|}{|S|} |S|(|F| - |S|)}.$$

Figure 2 shows the relationship of the weights and cardinality of the subsets. Note that the weights are symmetric and highest for "small" and "large" subsets. In order to reduce the computational complexity, the kernel SHAP algorithm subsamples the subsets. A fixed pre-determined threshold is used to restrict the number of subsets that will be sampled from all possible subsets. In practical implementations, this number is usually set to be $K = 2M + 2^{11}$. (Lundberg & Lee, 2017). It turns out, however, that a relatively small number of subsets is able to approximate the SHAP values very well. The Appendix provides results from a limited experimental study to support this assertion.

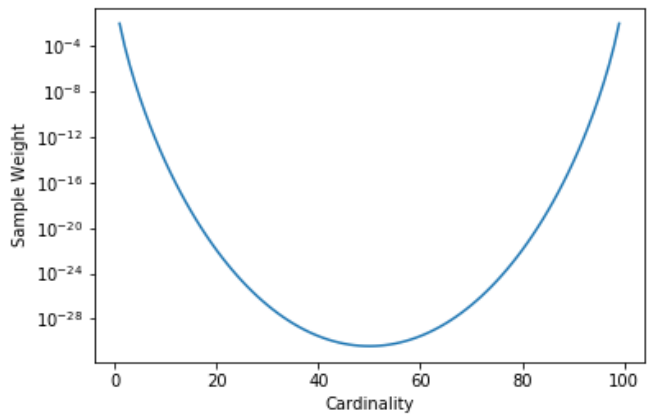

*Figure 2 Subset sample weight by cardinality of the subset for an example with 100 total features*

Another challenge associated with computing SHAP values is the difficulty in computing all the possible conditional expectations in equation (1). Lundberg and Lee (2017) proposed approximating them using marginal expectations. Specifically, $E[f(X)|X_S = x_S] = E_{X_{\bar{S}}|X_S}[f(X)]$ is approximated by $E_{X_{\bar{S}}}[f(X)]$, where $\bar{S}$ is the complementary set of $S$. Such an approximation will not yield a reasonable result in some situations. We also note that the dimension-reduction within in the CNN structure may ameliorate the computational burden somewhat.

### 3.3 CNN SHAP

#### 3.3.1. Method

To compute SHAP on a trained CNN model, we divide the CNN into two main parts: *feature extraction part* and *classification part* (see Figure 1). The feature extraction part is used to identify the important n-grams and outputs feature values. The classification part then looks at all extracted features and makes the prediction. To explain a particular prediction of the constructed CNN model, we take the following steps:

**Feature Extraction Step**: The word embedding vectorizes each word (and hence converts the unstructured data to structured data). But it adds to the explanation complexity since the embedding dimension is difficult to interpret. However, the convolution function applied over the word vectors transforms the vectors to a value that represents a particular meaning. Usually, words of similar meanings will be embedded into similar vectors (measured in terms of cosine similarity) and get close values from the same convolution function. Therefore, a filter should be capable of capturing a specific semantic meaning from the sentences. Figure 3 shows the structure of a convolutional filter and how it relates to the original bigram "are afraid".

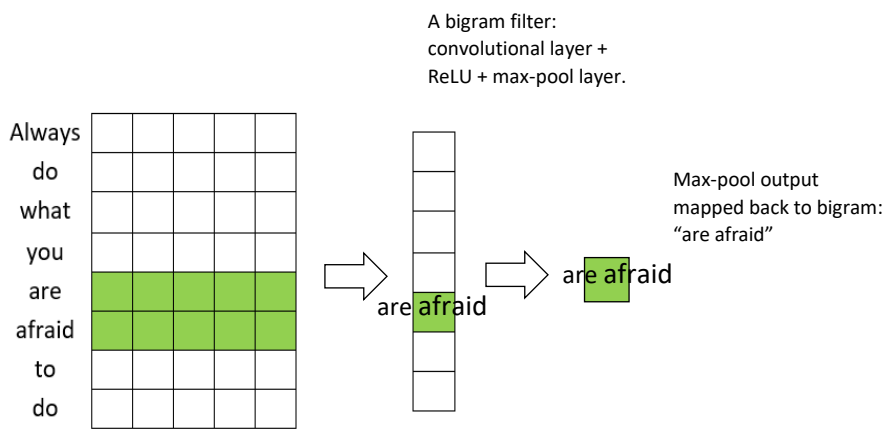

*Figure 3 Relate Convolutional Filter Output to Original Text N-gram. First, the embedded text documents of "Always do what you are afraid to do" is fed into a size 2 filter. The filter scans every bi-gram and uses the convolution + ReLU function to output a value for each scanned bi-gram respectively. These outputs form a size 7 vector. The filter then finds the maximum of the 7 values to be the final output in the end. In this picture, the output maximum value achieved by the convolution + ReLU output of bi-gram "are afraid".*

For an input text document of length $n$ represented by $x_i = \{x_{i1}, x_{i2}, \ldots, x_{in}\}$, by applying all $h$ filters, we will get a feature vector $\boldsymbol{\theta}_i = \{\theta_{i1}, \theta_{i2}, \ldots, \theta_{ih}\}$. Consequently, the unstructured texts for the entire dataset are transformed into structured data with $h$ features as follows:

$$\boldsymbol{\Theta} = \begin{pmatrix} \theta_{01} & \theta_{02} & \ldots & \theta_{0h} \\ \theta_{11} & \theta_{12} & \ldots & \theta_{1h} \\ \ldots & \ldots & \ldots & \ldots \\ \theta_{N1} & \theta_{N2} & \ldots & \theta_{Nh} \end{pmatrix},$$

where $N$ is the dataset size. Therefore, the problem of explaining unstructured text data becomes a problem of explaining this structured data.

**SHAP Explanation Step**: The feature extraction step reduces the feature dimension to $h$, which is usually a relatively small number in comparison to the vocabulary dimension. Now we can use kernel SHAP over the extracted features to get a good approximations to the exact SHAP values. In addition, we can also maintain the language structure to some extent. Originally, when applying kernel SHAP over vocabulary features, permutation is done by turning a particular word on and off. Suppose a word is not present in a particular text; turning it on could make a big difference to the final prediction. As a result, a nonexistent word in the text can receive high SHAP score. This can be avoided in CNN SHAP, where the permutation is done over features that are extracted by the filters. The SHAP score received by a filter is then attributed to the corresponding n-gram picked up by the filter.

**De-duplication Step**: Ideally in a well-trained CNN model, a filter is capable of catching a particular semantic feature, and different filters capture different semantic features. However, this is often not the case in practice. Duplication of filters in CNN is a prevalent phenomenon, and it increases with the number of filters in a convolutional layer (RoyChowdhury, Sharma, Learned-Miller, & Roy, 2017). Filters of similar sizes with high cosine similarity could result in selecting exactly the same n-grams from the text. Therefore, de-duplication of filters are necessary. Given two filters $f_i \in R^{n_1 \times m}, f_j \in R^{n_2 \times m}$, we define a $k$-partial cosine similarity as

$$s_k(f_i, f_j) = \max_{1 \leq k < \min(n_1, n_2)} \frac{f_i^{(k)} \cdot f_j^{(k)}}{||f_i^{(k)}|| \cdot ||f_j^{(k)}||},$$

where $f^{(k)} \in R^{k \times m}$ is a continuous sub-segment of filter $f$. One can plot the histograms of the $k$-partial cosine similarity calculated from all pairs of filters to investigate the filter similarities. It is also a reference of whether the de-duplication is necessary or not.

Furthermore, n-grams filtered by the CNN model may not ensure the semantic integrity. In some other occasions, the top n-grams are repeated and other important n-grams are pushed back. To solve those problems, we apply the following two steps: Exact De-duplication and Merge De-duplication. The summation of SHAP scores is used as the de-duplicated n-gram SHAP scores. At the exact de-duplication step, we delete the repeated n-grams and add up their SHAP scores. At the merge de-duplication step, all overlapped n-grams are merged together, and the SHAP scores are summed. For example, given a sentence "I called the customer service center", filter 1 receives score 0.500 for n-gram "called the customer", and filter 2 receives score 1.2 for "customer service", then the merge de-duplicated score will be 1.7 for the phrase "called the customer service". The final explanation is then achieved after both de-duplication steps.

Note that during the above de-duplication steps, we use summation of SHAP scores as the de-duplicated n-grams' SHAP score. This is because SHAP is an additive explanation model. Using summation retains the additivity of SHAP. Another option is maximization. In our experiments on the Amazon Elec dataset, we found that summation method outperforms maximization.

### 3.3.2 Real Data Study: Sentiment Analysis

We use the Amazon electronic products review (Amazon Elec) dataset (Johnson & Zhang, 2014) to illustrate the results. We used 50k training set and 25k testing set for our experiment. The constructed CNN model for sentiment analysis of the product reviews has:

1) An embedding layer with dimension of $1000 \times 100$, where 1000 is the padded sentence length, and 100 is the word embedding dimension. GloVe Wiki pre-trained embedding is used. The embedding layer is fixed during training process;
2) A convolutional layer with ReLU activation, where 1d convolutions in sizes $1 \times 100$, $2 \times 100$, and $3 \times 100$ are used. For each size, we use 50 convolutions. These convolutions correspond to unigrams, bi-grams, and tri-grams;
3) A max-pool layer; and
4) A linear layer consist of a linear layer with sigmoid activation.

Table 1 shows two examples of CNN SHAP results from Amazon Elec after the SHAP explanation step, where only the top 10 positive/negative n-grams are presented. Example 1 is a positive review and most of the filters receive positive SHAP scores. The top ranked n-grams are *no problems*, *is the excellent*, and *perhaps even better*, and all of them express highly positive attitudes. Example 2 is a negative review with filters receiving mostly negative SHAP scores. The top ranked n-grams are *be unusable*, *horrible* and *the bloody thing*.

| Amazon Elec Review Example 1 | | | Amazon Elec Review Example 2 | | |
| --- | --- | --- | --- | --- | --- |
| *"i have had my unit for almost 2 years and had no problems great unit but perhaps even better is the excellent service the service department recently replaced the rca adapter for free after i experienced some problems no questions asked that 's what i call service"* | | | *"horrible tv the first day we had it , the tv would get ' stuck ' on a digital channel it was trying to tune in and would be unusable for about 10 minutes to half an hour the second day , the bloody thing would n't even turn on either by remote or the power button on the tv shell out a few extra for a sony instead"* | | |
| Positive N-gram (Top 10) | | | Negative N-gram (Top 10) | | |
| Filter | N-gram | SHAP | Filter | N-gram | SHAP |
| 86 | no problems | 1.192 | 51 | be unusable | -0.517 |
| 135 | is the excellent | 0.834 | 45 | horrible | -0.423 |
| 108 | perhaps even better | 0.290 | 16 | unusable | -0.291 |
| 121 | i have had | 0.276 | 120 | the bloody thing | -0.276 |
| 120 | but perhaps even | 0.268 | 139 | trying to tune | -0.269 |
| 2 | great | 0.260 | 141 | a digital channel | -0.267 |
| 1 | rca | 0.248 | 126 | half an hour | -0.255 |
| 123 | is the excellent | 0.243 | 144 | extra for a | -0.238 |
| 136 | years and had | 0.218 | 60 | bloody thing | -0.232 |
| 35 | excellent | 0.201 | 93 | would be | -0.230 |

*Table 1 SHAP Explanation Step Results: Example 1 and Example 2 on Amazon Elec Review*

Table 2 is an example where the prediction of the sentiment is vague: there are highly scored positive n-grams as well as highly scored negative n-grams. Among the positive ones are *easy* and *great*, while the negative ones are like *worse* and *fail to make*. Additionally, we can see from Table 1 and Table 2 that filter 2 picks *great* in both example 1 and example 3, which expresses a positive attitude towards the electronic product. Filter 45 selects *horrible* in example 2 and *worse* in example 3, both of which show a negative attitude towards the products. This corresponds to our intuition that the filters trained in this CNN model tend to catch a specific type of semantic information across the all reviews. We also observe the filter duplication phenomenon. In example 3, filters 143 selects *will not connect*, 93 selects *will not*, which are overlapped n-grams. Similarly, filter 66 picks *'m waiting* and filter 145 picks *'m waiting for*.

We calculate the 1-partial cosine similarities for all pairs of filters' convolutional weights for Amazon Elec CNN. The distribution of the similarities is shown in Figure 4. We can see that the similarities of most of the filter pairs are relatively low, and only a few of them are higher than 0.6. This provides further evidence that there exist overlapped or duplicated n-grams picked by different filters, and the de-duplication step is necessary.

| CNN SHAP: Amazon Elec Review Example 3 | | | | | |
|---|---|---|---|---|---|
| *"right now i 'm waiting for a replacement hard drive this is the second time in 10 years i ordered it from <OOV> com for about 100 00 i will keep this unit working as long as possible as it has a lifetime tivo service contract associated with it it is easy to use , and pretty much foolproof i do n't use the modem others have complained about because ma bell has worse customer service than sony the unit will not connect with my internet phone service , and i now have failed to make a call for over 1600 days big deal they will never disconnect me it is not their policy i use this unit in a bedroom with a small 32 tv i know it 's not hdtv , but it produces a great picture as is i will keep this unit as long as possible"* | | | | | |
| Positive N-gram (Top 10) | | | Negative N-gram (Top 10) | | |
| Filter | N-gram | SHAP | Filter | N-gram | SHAP |
| 43 | easy | 0.728 | 45 | worse | -0.506 |
| 2 | great | 0.345 | 139 | failed to make | -0.401 |
| 118 | is easy to | 0.317 | 72 | worse customer | -0.398 |
| 81 | a bedroom | 0.296 | 28 | failed | -0.374 |
| 25 | complained | 0.289 | 143 | will not connect | -0.342 |
| 59 | Great picture | 0.284 | 93 | will not | -0.320 |
| 62 | i use | 0.263 | 55 | is not | -0.272 |
| 15 | easy | 0.252 | 66 | 'm waiting | -0.261 |
| 135 | is the second | 0.219 | 145 | 'm waiting for | -0.259 |
| 103 | has a lifetime | 0.219 | 86 | never disconnect | -0.206 |

*Table 2 Example 3 of CNN SHAP explanation on Amazon Elec Review*

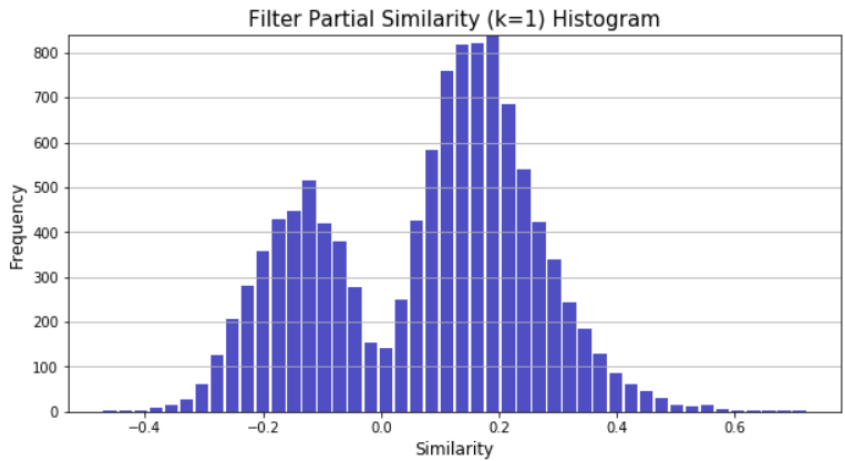

*Figure 4 Distribution of 1-partial filter similarities.*

Table 3 and Table 4 show the results after exact de-duplication and merge de-duplication are applied. In example 4, there are 3 repeated *extremely pleased* in top 10 n-grams. Actually, for all n-grams that receive positive SHAP scores, *extremely pleased* appears 4 times. We keep only 1 of them and add up all 4 SHAP scores. But problems still remain in some cases. For the same example 4, even after exact de-duplication, the top 3 n-grams are still highly overlapped: *extremely pleased*, *am extremely pleased*, and *extremely pleased with*. After we apply the merge de-duplication, all the overlapped n-grams (including one outside of top 10) are merged to *am extremely pleased with*, and their scores add up to 3.749.

### Global CNN SHAP: Amazon Elec Review Example 4

*"i am extremely pleased with this tv so i bought another one 55 with 3d inky blacks , i mean really black , as black as black can be beautiful colors easy set up with image science foundation approved calibration controls thx certified picture there are n't many tvs that are to thx certification standards for picture quality acurate reproduction of color the speakers are not very good the picture ease of set up are fantastic the remote is very good the tv looks great even when its not on , like a work of modern art ! buy it , you wo n't regret it"*

| Original Filter Results | | Deduplicated Results | | Merged Results | |
|---|---|---|---|---|---|
| N-gram | SHAP | N-gram | SHAP | N-gram | SHAP |
| am extremely pleased | 0.666 | extremely pleased | 1.482 | am extremely pleased with | 3.749 |
| extremely pleased | 0.651 | am extremely pleased | 1.236 | beautiful colors easy set up | 1.637 |
| am extremely pleased | 0.569 | extremely pleased with | 0.871 | is very good | 0.488 |
| extremely pleased with | 0.525 | is very good | 0.488 | science foundation approved | 0.483 |
| easy set up | 0.428 | easy set up | 0.428 | are fantastic | 0.375 |
| extremely pleased | 0.399 | are fantastic | 0.375 | looks great even | 0.320 |
| extremely pleased | 0.347 | beautiful colors easy | 0.345 | modern art ! buy | 0.270 |
| foundation approved | 0.298 | foundation approved | 0.298 | calibration controls thx certified | 0.228 |
| is very good | 0.292 | easy set | 0.271 | good the picture ease of | 0.183 |
| extremely pleased with | 0.288 | beautiful colors | 0.232 | acurate reproduction | 0.137 |

*Table 3 Deduping of Filter SHAP Values for Amazon Example 4.*

### Global CNN SHAP: Amazon Elec Review Example 5

*"i am very disappointed in this product i had previously purchased a <OOV> inch tv and was not pleased with that one , so i looked online for <OOV> brand and settled on the axion and was very dissatisfied with this also the picture was very erratic and would not stay with a good picture or sound i tried to adjust and could not get anything i would hope that others would not have to go through this thank you , <OOV> <OOV>"*

| Original Filter Results | | Deduplicated Results | | Merged Results | |
|---|---|---|---|---|---|
| N-gram | SHAP Score | N-gram | SHAP | N-gram | SHAP |
| very disappointed | -0.767 | very disappointed | -1.601 | am very disappointed in | -2.229 |
| tried to | -0.559 | tried to | -0.650 | erratic and would not stay with | -1.178 |
| very dissatisfied | -0.553 | very dissatisfied | -0.553 | sound i tried to adjust | -0.862 |
| very dissatisfied | -0.491 | disappointed | -0.431 | was very dissatisfied | -0.649 |
| very dissatisfied | -0.342 | would not stay | -0.299 | purchased a <OOV> inch tv | -0.343 |
| would not stay | -0.299 | stay | -0.285 | was not pleased with that | -0.340 |
| the picture was | -0.264 | the picture was | -0.264 | this product i had | -0.299 |
| for <OOV> brand | -0.166 | erratic | -0.228 | the picture was | -0.264 |
| stay | -0.165 | product | -0.211 | not get anything i | -0.258 |
| am very disappointed | -0.158 | for <OOV> brand | -0.166 | so i looked online | -0.189 |

*Table 4 Deduping of Filter SHAP Values for Amazon Example 5.*

## 4. Global CNN-Shapley Computations.

Lundberg and Lee (2017) also proposed summing up the modulus of SHAP scores across all samples to get a global SHAP score. These can however cause problems. For example, an n-gram may receive a very high SHAP score for one prediction (row) but low scores in the rest, or even in an extreme case it does not appear in any other predictions. Consequently, its total SHAP scores may be higher than other n-grams.

To address this, we propose a rank-based method for global explanation as follows:

1) For each n-gram $w$, consider its exact-deduplicated CNN-SHAP score for each observation (prediction) and compute its rank order by comparing its score with all other n-grams; retain it if it is within the top $k$, discard it otherwise. Denote the collected ranks for that n-gram as $R_w = \{r_w^{i_1}, r_w^{i_2}, \ldots, r_w^{i_m}\}$, where $r_w^i \in \{1, 2, \ldots, k\}$.
2) Include the n-gram in the important n-grams list if it ranks in the top $k$ for at least one document based on CNN SHAP.
3) Use one of the techniques below to calculate the global rank score for each n-gram in the important n-grams list.

There are several ways to calculate these combined rank scores. Among these, Inverse Rank Average (IRA) and Uniform Rank Average (URA) worked best in our experiments. The idea in IRA is to assign a decreasing function $s(\cdot)$ of the rank, and then compute an average for each set of ranks an n-gram receives:

$$s_{global}(R_w) = \frac{s\left(r_w^{i_1}\right) + s\left(r_w^{i_2}\right) + \cdots + s\left(r_w^{i_m}\right)}{m + k}.$$

The IRA score is defined as $s_{global}$ with $s(r) = \frac{1}{r}$, while the URA score is defined as $s_{global}$ with $s(r) = \frac{k+1-r}{k}$. Note that we use $m + k$ in the denominator to penalize the n-grams that hit the top $k$ less frequently. For example, an n-gram that has rank set of $\{1\}$ will receive lower rank score than an n-gram that has rank set of $\{1, 1, 2\}$. When we consider top 5 ranked n-grams, an n-gram with rank set of $\{1, 1, 1, 2, 2\}$ will receive the IRA score of

$$s_{IRA}(\{1, 1, 1, 2, 2\}) = \frac{(1 + 1 + 1 + 0.5 + 0.5)}{5 + 5} = 0.4$$

For the same example,

$$s_{URA}(\{1, 1, 1, 2, 2\}) = \frac{(1 + 1 + 1 + 0.8 + 0.8)}{5 + 5} = 0.46.$$

Once we compute the global scores for all n-grams in the important n-grams list, we can sort them by the global ranking scores provided above.

We illustrate the results on the Amazon Electronic Review dataset using the URA scores. The results are shown in Table 5. The positive n-grams are selected from all positive reviews, and the negative n-grams are selected from the negative reviews. Those N-grams that achieves top positive global scores express the positive sentiment, while those achieve top negative scores reveal negative attitudes towards the product. In particular, we can see the bi-gram *no problems* appears in top positive n-grams, while bi-gram *the problem* appears in top negative n-grams. This is a good example that bag-of-word explanation is problematic, while CNN SHAP is capable of catching sematic information and giving more meaningful interpretation.

| Global CNN SHAP scores based on URA: Amazon Elec Reviews | | | |
|---|---|---|---|
| Positive N-gram (Top 10) | | Negative N-gram (Top 10) | |
| N-gram | Score | N-gram | Score |
| easy | 0.836 | poor | 0.757 |
| very pleased with | 0.740 | waste | 0.754 |
| no problems | 0.727 | terrible | 0.733 |
| very happy with | 0.667 | not worth | 0.710 |
| i highly recommend | 0.663 | same problem | 0.687 |
| is a great | 0.625 | worst | 0.675 |
| the best | 0.614 | the problem | 0.689 |
| beat | 0.600 | does not work | 0.640 |
| love it | 0.599 | return it | 0.623 |
| i would recommend | 0.578 | refund | 0.620 |

*Table 5 Positive/Negative Top 10 N-grams of Amazon Elec Reviews picked up by Global CNN SHAP.*

## 5. Conclusions and Future Work

We have developed a novel approach for computing SHAP scores for CNN-based text classification models with NLP data. CNN SHAP fully utilizes the advantage of SHAP explanation in computing local feature importance, but avoids the curse of high feature dimensionality in explaining NLP tasks. By applying de-duplication in CNN SHAP output, we are able to provide more understandable and meaningful explanations of the model. We also proposed global explanation based on CNN SHAP, which quantifies the importance of n-grams globally.

Another potential usage of CNN SHAP is as a surrogate explanation method approximating any text classification models, especially those much more complicated DNN models, using CNN and using the techniques here to explain the results. However, more work needs to be done to evaluate the performance of CNN SHAP as a surrogate explanation method. In addition, the method uses marginal expectations to ease the computational burdens in evaluating the conditional expectations in the SHAP formula in equation (1). This can perform poorly in some applications, so it in necessary to explore alternatives.

## Appendix A

In order to quantify how much information we might lose by sampling the subsets under a fixed threshold, we conducted a simulation study. Define the *weight capture rate* as the total weight of sampled subsets over the total weight of all possible subsets. This helps us to learn the impact of the number of subsets chosen as the feature size of $M$ is getting larger. Figure 5 shows the captured weight in KS approximation when sample threshold is set to 1,000. It can be easily seen that the weight capture rate drops dramatically as the feature dimension increases. When the feature dimension exceeds 1000, the captured weight is as low as 0.066.

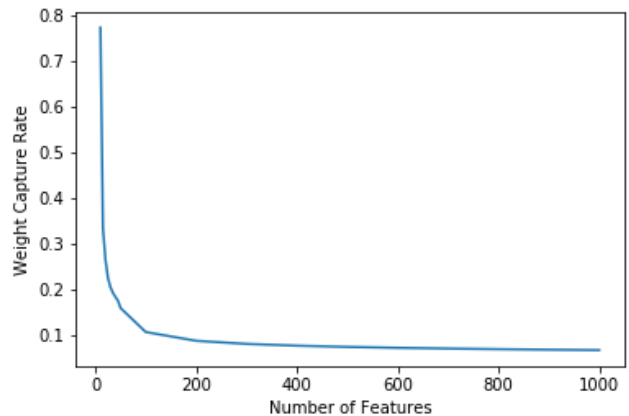

*Figure 5 Weight capture rate when different number of features are used in the prediction model. The sample threshold is 1000.*

We use the following simulation that even with 1/10 of total subsets used, which is equivalent to 1/3 of total weight captured, KS is still able to approximate SHAP values very well. Let

$$f(X) = 1.5X_1 + 1.5X_2 + 1.5X_3 + X_4 + 1.4X_5 + 0.5X_6 + 1.8X_7 + 1.8X_8 - X_9 - 1.5X_{10},$$

where $X = (X_1, X_2, \ldots, X_{10})^T \sim MultivariateNomal(\boldsymbol{\mu}, \boldsymbol{\Sigma})$. We set $\boldsymbol{\mu} = (0, 0, \ldots, 0)^T$, and

$$\boldsymbol{\Sigma} = \begin{pmatrix} 1.0 & 0.1 & 0.0 & 0.0 & 0.0 & 0.0 & 0.0 & 0.0 & 0.0 & 0.0 \\ 0.1 & 1.0 & 0.0 & 0.0 & 0.0 & 0.0 & 0.0 & 0.0 & 0.0 & 0.0 \\ 0.0 & 0.0 & 1.0 & 0.1 & 0.1 & 0.0 & 0.0 & 0.0 & 0.0 & 0.0 \\ 0.0 & 0.0 & 0.1 & 1.0 & 0.1 & 0.0 & 0.0 & 0.0 & 0.0 & 0.0 \\ 0.0 & 0.0 & 0.1 & 0.1 & 1.0 & 0.0 & 0.0 & 0.0 & 0.0 & 0.0 \\ 0.0 & 0.0 & 0.0 & 0.0 & 0.0 & 1.0 & 0.1 & 0.1 & 0.1 & 0.1 \\ 0.0 & 0.0 & 0.0 & 0.0 & 0.0 & 0.1 & 1.0 & 0.1 & 0.1 & 0.1 \\ 0.0 & 0.0 & 0.0 & 0.0 & 0.0 & 0.1 & 0.1 & 1.0 & 0.1 & 0.1 \\ 0.0 & 0.0 & 0.0 & 0.0 & 0.0 & 0.1 & 0.1 & 0.1 & 1.0 & 0.1 \\ 0.0 & 0.0 & 0.0 & 0.0 & 0.0 & 0.1 & 0.1 & 0.1 & 0.1 & 1.0 \end{pmatrix}.$$

The error term is added when we generate the $i$-th response:

$$y_i = f(\boldsymbol{X}_{(i)}) + \epsilon_i,$$

where $\epsilon_i \sim Normal(0, 1)$.

Then, with sample threshold of 100, 200, and 500, which is approximately 0.1, 0.2, and 0.49 of total subsets, we have the weight capture rate equal to 0.34, 0.40, and 0.49. As shown in Table 6, KS does a good job of estimating the true SHAP values even we use only 100 subsets out of $2^{10}$ total subsets. When we use 500 subsets, the differences from true SHAP values are negligible.

| X | Example 1 | | | | |True - KS| for 100 subsets | | |True - KS| for 200 subsets | | |True - KS| for 500 subsets | |
|---|---|---|---|---|---|---|---|---|---|---|
| | TRUE | KS, 100 subsets | KS, 200 subsets | KS, 500 subsets | Mean | Std. | Mean | Std. | Mean | Std. |
| x1 | 0.2892 | 0.2738 | 0.2825 | 0.2894 | 0.0488 | 0.113 | 0.0192 | 0.0542 | 2.11E-04 | 1.17E-14 |
| x2 | 0.8685 | 0.8698 | 0.8615 | 0.8681 | 0.0478 | 0.1119 | 0.0175 | 0.049 | 3.70E-04 | 1.17E-14 |
| x3 | 0.2866 | 0.2813 | 0.2771 | 0.2861 | 0.0471 | 0.1098 | 0.0173 | 0.0508 | 4.57E-04 | 1.18E-14 |
| x4 | 0.9963 | 0.9901 | 0.9913 | 0.9964 | 0.0537 | 0.1243 | 0.0209 | 0.0565 | 4.64E-05 | 8.63E-15 |
| x5 | -0.0584 | 0 | 0 | -0.0586 | 0.0519 | 0.1243 | 0.0204 | 0.063 | 2.54E-04 | 1.09E-14 |
| x6 | -1.383 | -1.3909 | -1.3909 | -1.3831 | 0.0714 | 0.1357 | 0.0291 | 0.0704 | 1.71E-04 | 5.57E-15 |
| x7 | 1.8455 | 1.8442 | 1.838 | 1.8457 | 0.0426 | 0.102 | 0.0169 | 0.0561 | 1.46E-04 | 1.38E-14 |
| x8 | 0.407 | 0.3883 | 0.4052 | 0.4067 | 0.046 | 0.1077 | 0.0164 | 0.0463 | 2.67E-04 | 1.42E-14 |
| x9 | 1.2478 | 1.2417 | 1.2418 | 1.2478 | 0.0564 | 0.1301 | 0.0207 | 0.0574 | 7.21E-05 | 8.71E-15 |
| x10 | 2.298 | 2.2909 | 2.2899 | 2.298 | 0.0528 | 0.1309 | 0.0198 | 0.0596 | 4.93E-05 | 9.10E-15 |

*Table 6 Simulation results of kernel SHAP Approximation of the true SHAP values.*